\documentclass[conference, a4paper]{IEEEtran}
\IEEEoverridecommandlockouts
\usepackage{cite}
\usepackage{amsmath,amssymb,amsfonts}
\usepackage{algorithmic}
\usepackage{graphicx}
\usepackage{textcomp}
\usepackage{xcolor}
\usepackage{booktabs}
\usepackage{amssymb}
\usepackage{makecell}
\usepackage{multirow}
\def\BibTeX{{\rm B\kern-.05em{\sc i\kern-.025em b}\kern-.08em
    T\kern-.1667em\lower.7ex\hbox{E}\kern-.125emX}}
\begin{document}

\title{Contour Information Aware 2D Gaussian Splatting for Image Representation}

\makeatletter
\newcommand{\linebreakand}{%
  \end{@IEEEauthorhalign}
  \hfill\mbox{}\par
  \mbox{}\hfill\begin{@IEEEauthorhalign}
}
\makeatother

\author{
\IEEEauthorblockN{Masaya Takabe}
 \IEEEauthorblockA{
    \textit{Graduate School of FSE,}\\
    \textit{Waseda University} \\
    Tokyo, Japan \\
    masaya.ta@asagi.waseda.jp}

\and
\IEEEauthorblockN{Hiroshi Watanabe}
  \IEEEauthorblockA{
    \textit{Graduate School of FSE,} \\
    \textit{Waseda University} \\
    Tokyo, Japan \\
    hiroshi.watanabe@waseda.jp}
\and
\IEEEauthorblockN{Sujun Hong}
 \IEEEauthorblockA{
    \textit{Sharp Corporation} \\
    Chiba, Japan \\
    hong.sujun@mail.sharp}
\and
\IEEEauthorblockN{Tomohiro Ikai}
 \IEEEauthorblockA{
    \textit{Sharp Corporation} \\
    Chiba, Japan \\
    ikai.tomohiro@mail.sharp}
\linebreakand
\IEEEauthorblockN{Zheming Fan}
 \IEEEauthorblockA{
    \textit{Sharp Corporation} \\
    Chiba, Japan \\
    fan.zheming@mail.sharp}
\and
\IEEEauthorblockN{Ryo Ishimoto}
 \IEEEauthorblockA{
    \textit{Sharp Corporation} \\
    Chiba, Japan \\
    ishimoto.ryo@mail.sharp}
\and
\IEEEauthorblockN{Kakeru Sugimoto}
 \IEEEauthorblockA{
    \textit{Sharp Corporation} \\
    Chiba, Japan \\
    sugimoto.kakeru@mail.sharp}
\and
\IEEEauthorblockN{Ruri Imichi}
 \IEEEauthorblockA{
    \textit{Sharp Corporation} \\
    Chiba, Japan \\
    imichi.ruri@mail.sharp}
}

\maketitle

\begin{abstract}
    Image representation is a fundamental task in computer vision.
    Recently, Gaussian Splatting has emerged as an efficient representation framework, and its extension to 2D image representation enables lightweight, yet expressive modeling of visual content.
    While recent 2D Gaussian Splatting (2DGS) approaches provide compact storage and real-time decoding, they often produce blurry or indistinct boundaries when the number of Gaussians is small due to the lack of contour awareness.
    In this work, we propose a Contour Information-Aware 2D Gaussian Splatting framework that incorporates object segmentation priors into Gaussian-based image representation.
    By constraining each Gaussian to a specific segmentation region during rasterization, our method prevents cross-boundary blending and preserves edge structures under high compression.
    We also introduce a warm-up scheme to stabilize training and improve convergence.
    Experiments on synthetic color charts and the DAVIS dataset demonstrate that our approach achieves higher reconstruction quality around object edges compared to existing 2DGS methods.
    The improvement is particularly evident in scenarios with very few Gaussians, while our method still maintains fast rendering and low memory usage.
\end{abstract}


\begin{figure*}[h]
    \centering
    \includegraphics[width=\textwidth]{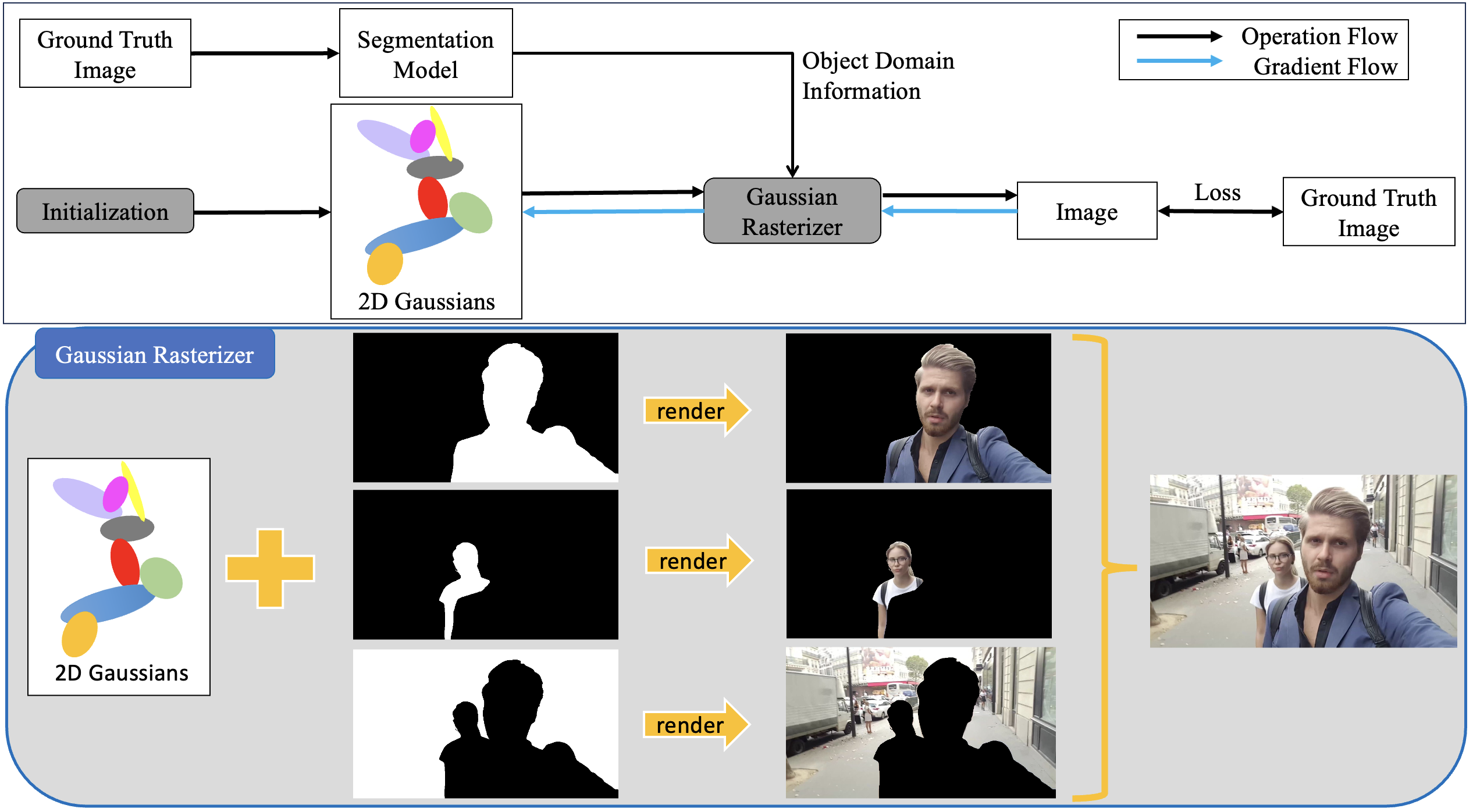}
    \caption{Overview of the proposed method.
        Top: The pipeline of our framework.
        In addition to the conventional 2DGS pipeline, we feed the target image (to be represented with 2D Gaussians) into a segmentation model to obtain object-wise segmentation masks.
        These masks are integrated into the Gaussian Rasterizer as Object Domain Information.
        Bottom: Details inside the Gaussian Rasterizer.
        Each segmentation mask is represented using only the 2D Gaussians that fall within its region.
    }
    \label{fig:overview}
\end{figure*}

\section{Introduction}
    Image representation has long been a central problem in computer vision and graphics.
    With the rise of deep learning, Implicit Neural Representations (INRs) have emerged as a powerful paradigm for encoding images and scenes in a continuous and compact manner.
    By parameterizing visual signals with neural networks, INRs enable resolution-independent rendering and flexible content representation.
    However, INR-based methods often rely on large multilayer perceptrons (MLPs) or high-dimensional feature grids, which result in heavy memory usage and slow inference, limiting their practicality for interactive or resource-constrained applications.

    To address these challenges, recent research has explored alternatives based on Gaussian Splatting.
    In particular, 2D Gaussian Splatting (2DGS) methods such as Image-GS~\cite{zhang2025image} and GaussianImage~\cite{zhang2024gaussianimage} represent images with adaptive sets of Gaussian primitives, achieving compact storage and real-time decoding while retaining competitive visual quality.
    These methods completely avoid large networks, making them highly efficient and interpretable compared to traditional INRs.
    Despite their advantages, existing 2DGS techniques struggle in scenarios with extreme compression, i.e., when only a very limited number of Gaussians are available.
    Under such conditions, object boundaries often become blurry or indistinct because there is a lack of explicit incorporation of structural and contour information to guide Gaussian placement and blending.
    
    To overcome this limitation, we propose a novel image representation framework that incorporates contour priors into 2DGS.
    Our method leverages segmentation masks obtained from pre-trained models such as SAM to assign region-based constraints to each Gaussian during training and rendering.
    Specifically, we associate each Gaussian with a segmentation region and restrict its contribution to pixels within that region, thereby preserving structural consistency.
    This design enables more accurate reconstruction around object boundaries, even under a limited Gaussian budget with very few Gaussians.
    
    Through experiments on synthetic and real-world images, we demonstrate that the proposed method achieves higher reconstruction quality, particularly around edges, compared to conventional 2DGS under the same Gaussian budget.
    Furthermore, our approach remains compatible with standard 2DGS rasterizers and retains the advantages of fast rendering speed and low memory footprint.

    Our contributions are summarized as follows:
    \begin{itemize}
        \item We propose a contour-aware 2D Gaussian Splatting framework for efficient image representation, which improves edge fidelity under a limited Gaussian budget.
        \item We introduce a region-constrained rasterization strategy with a warm-up scheme to enforce contour consistency and stabilize training.
        \item We validate our method on synthetic color charts and real-world images, demonstrating superior reconstruction performance while retaining fast rendering and low memory usage.
    \end{itemize}

\section{Related Work}
\subsection{Implicit Neural Representation}
    Implicit Neural Representations (INRs) have become a versatile tool for encoding visual and signal data, enabling continuous and compact parameterizations across diverse domains.
    With the advent of deep learning, INRs have been widely applied to diverse tasks such as 3D scene representation~\cite{mildenhall2021nerf, sitzmann2020implicit}, 2D image representation~\cite{tancik2020fourier, dupont2021coin} and video representation~\cite{chen2021nerv}, super-resolution~\cite{chen2021learning}, and compression~\cite{dupont2022coin++}.
    A common class of INR methods relies on multilayer perceptrons (MLP) to map input spatial coordinates to color values, often using positional encodings~\cite{tancik2020fourier} or periodic activation functions~\cite{sitzmann2020implicit} to improve high-frequency detail modeling.
    Although these approaches achieve impressive reconstruction quality, they typically suffer from long training times, high GPU memory consumption, and slow decoding speeds, especially when applied to high-resolution content.
    To address these issues, more efficient variants have been proposed.
    However, even optimized methods remain computationally demanding for real-time or resource-constrained scenarios due to heavy memory access and limited content adaptivity.
    In contrast, our work explores an alternative framework that eliminates the reliance on large MLPs, instead leveraging 2DGS for fast, efficient, and interpretable image encoding.

\subsection{Gaussian Splatting for Visual Representation}
    Gaussian Splatting has recently emerged as a highly efficient and expressive method for 3D scene representation~\cite{kerbl20233d}, and has been extended to segmentation~\cite{zhou2024feature, cen2025segment}, dynamic scenes~\cite{li2024spacetime, sun2024splatter}, and generative modeling~\cite{zhou2024diffgs, yi2024gaussiandreamer}.
    Encouraged by its success in 3D vision, recent works have begun adapting Gaussian Splatting to 2D image representation.
    Recent works have explored Gaussian Splatting as a lightweight alternative for image representation.
    In this framework, an image $I(x,y)$ is approximated by a weighted sum of 2D anisotropic Gaussians:
    \begin{equation}
        I(x,y) = \sum_{i=1}^N \alpha_i \cdot \mathcal{N}\!\left((x,y); \mu_i, \Sigma_i \right) \cdot \mathbf{c}_i ,
    \end{equation}
    where each Gaussian primitive is parameterized by its center $\mu_i$, covariance $\Sigma_i$, opacity $\alpha_i$, and color $\mathbf{c}_i$.
    Unlike 3D Gaussian Splatting~\cite{kerbl20233d}, which requires depth-sorted $\alpha$-blending, 2DGS does not involve ordering along the depth dimension and therefore adopts this additive formulation as standard practice~\cite{zhang2024gaussianimage,zhang2025image}.
    
    Building on this idea, Image-GS\cite{zhang2025image} places Gaussians according to image gradient magnitudes, assigning more primitives to high-frequency regions such as edges and textures.
    GaussianImage\cite{zhang2024gaussianimage} instead proposes a fixed view 2DGS framework and introduces a rasterizer that accumulates Gaussian contributions through summation, rather than traditional alpha blending.
    Beyond static images, several works have also begun extending 2DGS to video representation\cite{wang2025gsvc}, demonstrating its potential for efficient spatio-temporal encoding.
    These approaches highlight that 2DGS offers compact storage and efficient rendering competitive with neural implicit representations.
    However, existing methods struggle to preserve sharp contours under extreme compression, where the number of Gaussians is severely limited.
    
    Our work builds upon this foundation by introducing contour-aware masking and Gaussian region control during rendering.
    This improves edge fidelity without sacrificing compression performance, bridging the gap between semantic structure preservation and efficient image encoding.

\begin{figure*}[h]
    \centering
    \includegraphics[width=\textwidth]{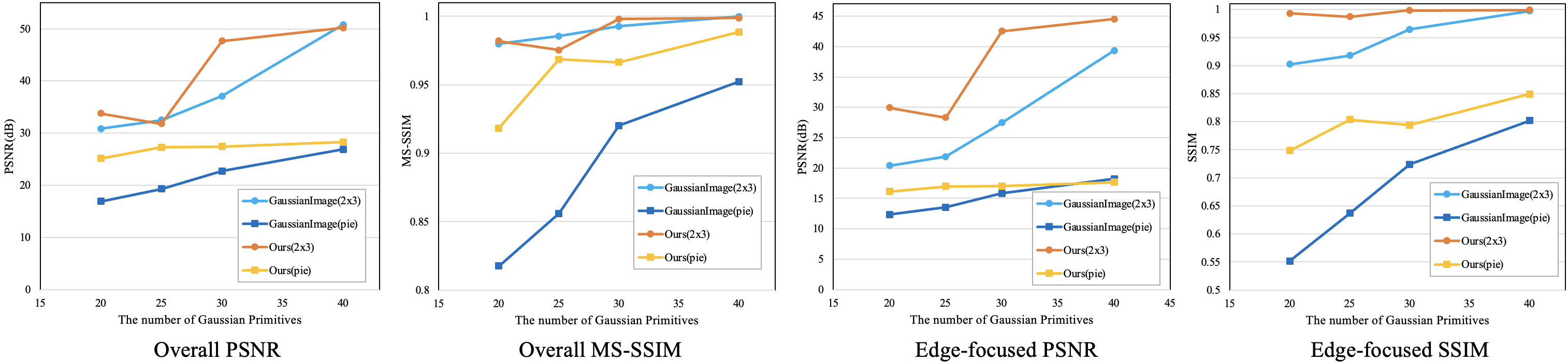}
    \caption{
        Quantitative results on the synthetic color charts.
        From left to right: PSNR and MS-SSIM over the entire image, PSNR and SSIM around edge regions.}
    \label{fig:color_chart_result_graph}
\end{figure*}

\begin{figure*}[h]
    \centering
    \includegraphics[width=\textwidth]{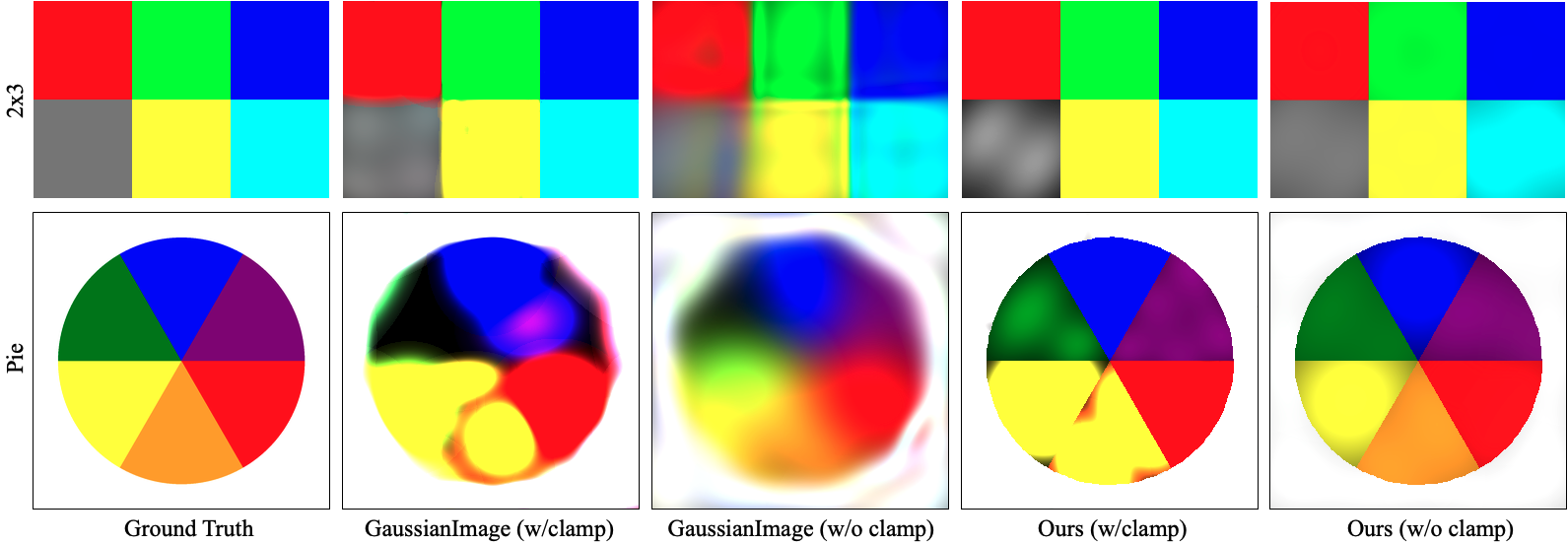}
    \caption{
        Qualitative results on the synthetic color chart with high compression, where the number of Gaussians is 20.
        From left to right: ground truth, baseline with clamping, baseline without clamping, ours with clamping, and ours without clamping.}
    \label{fig:color_chart_result_image}
\end{figure*}

\section{Proposed Method}
\subsection{Contour Information-Aware 2D Gaussian Splatting}
    To improve the preservation of object boundaries under limited Gaussian capacity, we propose a segmentation-guided 2DGS framework that explicitly incorporates semantic region constraints during rendering.
    Let $\mathcal{I}_{\text{gt}} \in \mathbb{R}^{H \times W \times 3}$ denote the ground truth image, and $\mathcal{M} \in \{1, \ldots, R\}^{H \times W}$ be a segmentation mask that assigns each pixel $(x, y)$ to one of $R$ regions.
    These masks are obtained using an existing segmentation model such as SAM~\cite{kirillov2023segment, ravi2024sam}, DINO\cite{caron2021emerging}, and LSeg~\cite{li2022language}.
    Each 2D Gaussian $g_i$ is parameterized by a tuple:
    \begin{equation}
        g_i = \left( \mu_i, \Sigma_i, \mathbf{c}_i, \alpha_i, r_i \right),
    \end{equation}
    where $\mu_i$ and $\Sigma_i$ denote the 2D mean and covariance matrix, $\mathbf{c}_i$ is the RGB color vector, $\alpha_i$ is the opacity, and $r_i \in \{1, \ldots, R\}$ is the assigned region ID.Importantly, $r_i$ is not a learnable parameter, but a fixed assignment derived from the segmentation mask.
    For rendering, we assign a region ID $r_{xy}$ to each pixel $(x, y)$ using the segmentation map: $r_{xy} = \mathcal{M}(x, y)$.
    The final rendered image $\mathcal{I}_{\text{pred}} \in \mathbb{R}^{H \times W \times 3}$ is then computed as:
    \begin{equation}
        \mathcal{I}_{\text{pred}}(x, y) = \sum_{i : r_i = r_{xy}} \alpha_i \cdot \mathcal{N}\left( (x, y); \mu_i, \Sigma_i \right) \cdot \mathbf{c}_i,
    \end{equation}
    where $\mathcal{N}\left( (x, y); \mu_i, \Sigma_i \right)$ is the Gaussian kernel centered at $\mu_i$ with covariance $\Sigma_i$, evaluated at the pixel $(x, y)$.
    This region-aware masking enforces that each Gaussian only contributes to the pixels within its assigned semantic region, preventing boundary artifacts and improving edge fidelity.
    Crucially, the masking is implemented using integer ID comparisons (i.e., $r_i = r_{xy}$), which ensures that rendering remains parallelizable and GPU-efficient.
    We do not modify the loss function, which remains the standard image-reconstruction loss:
    \begin{equation}
        \mathcal{L} = \| \mathcal{I}_{\text{gt}} - \mathcal{I}_{\text{pred}} \|_1,
    \end{equation}
    allowing optimization to focus on global reconstruction quality, while the mask constraints encourage more accurate local (edge) reconstruction.
    Fig.~\ref{fig:overview} provides an overview of the proposed region-constrained rendering pipeline.

\subsection{Region Assignment of Gaussians}
    The region ID $r_i$ of each Gaussian is determined by referencing its 2D position $\mu_i$ in the segmentation mask $\mathcal{M}$.
    Specifically, at initialization, we assign:
    \begin{equation}
        r_i = \mathcal{M}\!\left( \lfloor \mu_i^x \rfloor, \lfloor \mu_i^y \rfloor \right),
    \end{equation}
    where $(\mu_i^x, \mu_i^y)$ denotes the center position of Gaussian $g_i$.
    To account for positional updates of Gaussians during optimization, we introduce a \textit{warm-up strategy} for region ID assignment.
    In practice, $r_i$ is refreshed every 1000 iterations until half of the total training iterations (warm-up stage).
    After this stage, $r_i$ is fixed, stabilizing the training process while ensuring that Gaussians near the boundary of the object are correctly assigned in the early learning phase.

\subsection{Clamping-Free Training}
    In conventional 2DGS, pixel values are often clamped to the range [0, 1] during training.
    This clamping stabilizes the optimization process and accelerates convergence, since it avoids extreme pixel values in early iterations.
    However, clamping also restricts the expressive ability of Gaussians, especially in terms of intermediate color representation, such as dark green, orange, purple, or cyan.
    Intermediate colors are usually obtained by combining multiple Gaussians with different color weights.
    Therefore, when clamping is enforced, a larger number of Gaussians are required to reproduce such colors faithfully.
    In our design, we remove the clamping operation.
    Although this could normally lead to blurred boundaries when only a few Gaussians are used, our region-guided rasterization compensates for this drawback.
    Because object boundary information is explicitly introduced via segmentation masks, sharp contours can still be preserved even without clamping.
    Consequently, our method prioritizes the accurate reproduction of intermediate colors while maintaining contour fidelity, which is especially beneficial under limited Gaussian budgets.

\section{Experiments}
\subsection{Experimental Setup}
    We conduct both synthetic and real-world experiments to evaluate the effectiveness of our proposed segmentation mask-guided 2DGS framework.
    For synthetic tests, we use two types of color chart images: a $2\times3$ chart (200$\times$300 resolution) and a pie chart (300$\times$300).
    The $2\times3$ chart consists only of straight-line boundaries between color blocks, making it a highly ideal case for our method, which explicitly preserves object contours.
    In contrast, the pie chart contains curved and slightly blurred boundaries when zoomed in, providing a challenging scenario to test the robustness of edge representation.
    For real-world evaluation, we employ the DAVIS dataset~\cite{perazzi2016benchmark, pont20172017}, which was originally designed for video object segmentation, but contains high-quality, pixel-level annotated segmentation masks.
    We utilize the first frame of each video sequence to form a set of static test images.
    In this experiment, we assume access to ground truth segmentation masks and directly use them during training to validate whether our method functions effectively under almost perfect domain guidance.
    This setup allows us to isolate and test the core effectiveness of our framework, without being affected by segmentation model accuracy.
    All models are trained with an NVIDIA RTX A6000 GPU.
    Rotation and scaling matrices of the 2D Gaussians are efficiently computed via Cholesky decomposition for numerical stability and speed.
    Evaluation metrics include the peak signal-to-noise ratio (PSNR) and the Structural Similarity index (SSIM).
    In addition, to quantify the effectiveness of contour preservation, we also perform an edge-focused evaluation.
    Specifically, we extract a band of 5 pixels inward and outward from the ground truth segmentation boundary and compute the reconstruction metrics only within this narrow edge region.
    We refer to the PSNR measured in this band as Edge-focused PSNR (EF-PSNR).
    For edge-focused evaluation, we use SSIM instead of MS-SSIM, as the latter can be less stable on a narrow edge region.

\begin{figure*}[h]
    \centering
    \includegraphics[width=\textwidth]{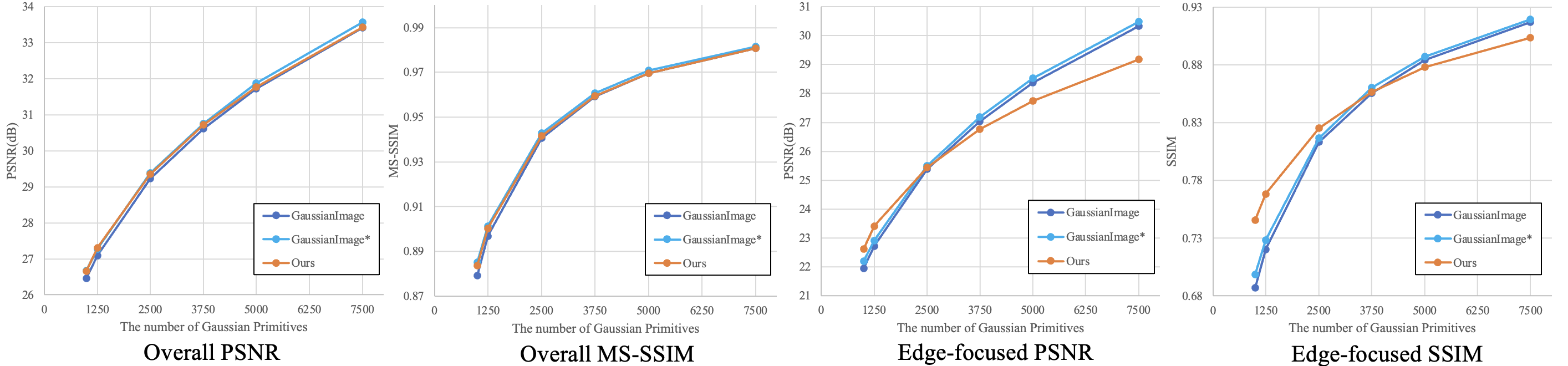}
    \caption{
        Quantitative results on the DAVIS dataset.
        From left to right: PSNR and MS-SSIM over the entire image, PSNR and SSIM around edge regions.
        Here, GaussianImage refers to the original method with clamping, while GaussianImage${}^*$ indicates its variant without clamping.
        }
    \label{fig:davis_result_graph}
\end{figure*}

\begin{figure*}[h]
    \centering
    \includegraphics[width=\textwidth]{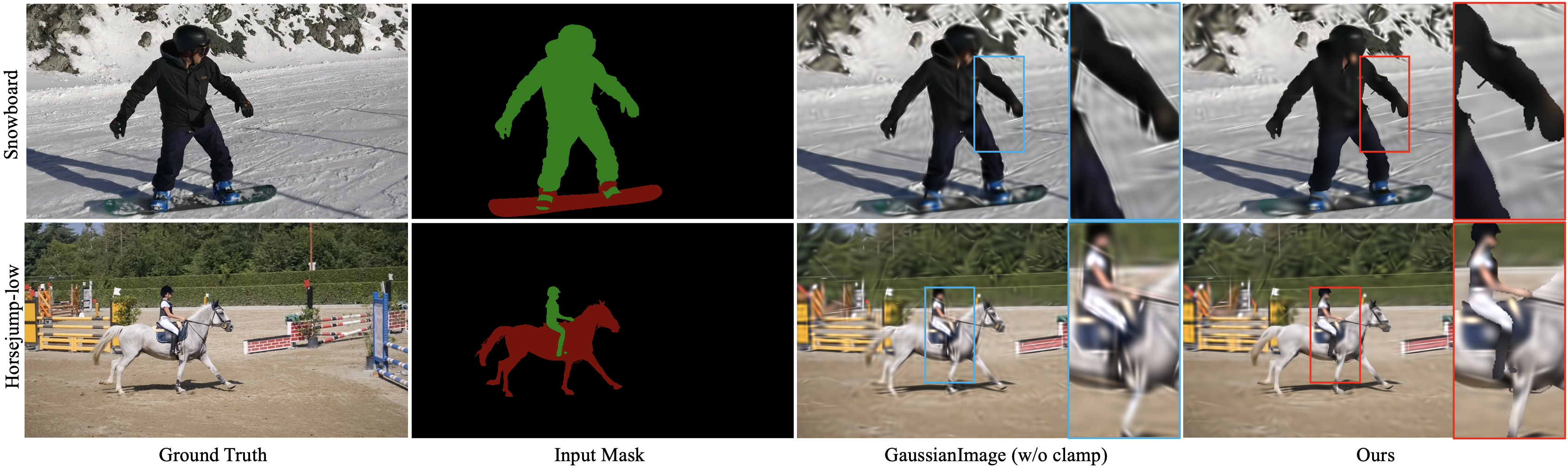}
    \caption{
        Qualitative results on the DAVIS dataset with high compression, where the number of Gaussians is 1250.
        From left to right: ground truth, input mask for the rasterizer (object domain information), baseline without clamping, and ours.}
    \label{fig:davis_result_image}
\end{figure*}

\subsection{Color Chart Results}
    Fig.~\ref{fig:color_chart_result_graph} summarizes the reconstruction performance in synthetic color charts.
    Under a constrained number of Gaussians (20 Gaussians, 50,000 training iterations), our method consistently outperforms baseline 2DGS in both PSNR and SSIM, especially within the edge regions.
    In the 2 × 3 chart, where the boundary lines are straight and well defined, our contour information-guided approach is highly effective in preserving the sharpness of edges.
    In contrast, while both methods struggle more with the curved and partially blurred transitions in the pie chart, our method still shows superior localization of boundaries and better perceptual quality.
    These results indicate that explicit incorporation of contour information not only improves reconstruction accuracy but also enhances structural consistency in challenging compression settings.
    
    Fig.~\ref{fig:color_chart_result_image} further provides qualitative comparisons under different clamping strategies.
    We consider four conditions: 1) baseline without clamping, 2) baseline with clamping, 3) our method with clamping, and 4) our method without clamping.
    Without clamping, the baseline can reproduce intermediate colors, i.e., colors where the RGB channels take values other than 0 or 1 (such as dark green, orange, and purple), but suffers from blurry contours.
    With clamping, the baseline produces sharper boundaries but fails to represent intermediate colors since only extreme RGB values dominate the blending.
    In contrast, our method without clamping successfully reproduces intermediate colors, while simultaneously maintaining sharp contours thanks to region-guided rasterization.
    This demonstrates that removing clamping, when combined with region guidance, achieves the best trade-off between color expressiveness and boundary fidelity.

\subsection{Real-world Dataset Evaluation}
    We present the quantitative results using the DAIVS dataset in Fig.~\ref{fig:davis_result_graph}.
    When representing images with the few number of Gaussians, our proposed method outperforms the baseline across all evaluation metrics, demonstrating the effectiveness of providing masks as region guidance.
    Moreover, Fig.~\ref{fig:davis_result_image} shows the corresponding reconstructed images.
    Compared to the baseline, our method yields clearer boundaries along the masked regions and produces cleaner backgrounds as a result of the improved boundary representation.
    On the other hand, when the number of Gaussians increases, our method sometimes performs worse than the baseline.
    This is because with a larger Gaussian budget, the Gaussians can fit the target image more accurately, even beyond the boundaries specified by the masks.
    Therefore, the effectiveness of our method is most evident in cases where the number of Gaussians is insufficient to accurately represent the region boundaries.
    
    In this experiment, the number of segmented regions was much smaller than that of typical segmentation models.
    We therefore expect this trend to become more pronounced as the number of segmented regions increases: our method tends to gain more advantage under high-compression (few Gaussians) settings, while less under low-compression (many Gaussians) settings.

\subsection{Ablation Study}
    To validate the contribution of each component in our framework, we conduct an ablation study on the real-world DAVIS dataset.
    Following the same setting as in the main experiments, we report results under two different Gaussian budget conditions: \textit{Few Gaussians} with 1250 primitives and \textit{Many Gaussians} with 7500 primitives.
    The quantitative results are summarized in Table~\ref{tab:ablation}.
    In addition to the standard PSNR metric, we also report EF-PSNR, which emphasizes reconstruction quality around object boundaries.
    We first observe that removing clamping improves PSNR in both settings, particularly in the many Gaussians case where the baseline reaches the highest value.
    However, this comes at the cost of blurrier contours, which is reflected in the lower edge-focused PSNR.
    In contrast, when combined with contour guidance, the removal of the clamping successfully reproduces intermediate colors while maintaining sharper boundaries, which is consistent with the qualitative results shown in Fig.~\ref{fig:color_chart_result_image}.
    Contour guidance alone brings limited benefits, especially under the few Gaussians condition.
    However, when coupled with the warm-up strategy, it leads to a clear improvement in edge-focused PSNR, demonstrating that warm-up is essential for stabilizing edge-aware optimization.
    Finally, the full model that integrates contour guidance, warm-up, and clamp removal achieves the best overall performance.
    It yields the highest scores in the few Gaussians conditions for both PSNR and EF-PSNR, highlighting its effectiveness under constrained Gaussian budgets.
    Although the improvement is less pronounced in the many Gaussians setting, where sufficient capacity already alleviates edge degradation, our design still delivers competitive PSNR while maintaining strong structural consistency.
    These results confirm the validity of our proposed design and the complementary roles of each component.

\begin{table}[t]
  \centering
  \caption{Ablation study of different components}
  \label{tab:ablation}
  \begin{tabular}{
    >{\centering\arraybackslash}p{7mm} 
    >{\centering\arraybackslash}p{4mm} 
    >{\centering\arraybackslash}p{8mm}|
    >{\centering\arraybackslash}p{7mm}
    >{\centering\arraybackslash}p{13mm}|
    >{\centering\arraybackslash}p{7mm}
    >{\centering\arraybackslash}p{13mm}}
    \toprule
    \multirow{2}{*}{\makecell{Contour \\ Guidance}} & 
    \multirow{2}{*}{\makecell{Warm \\ Up}} & 
    \multirow{2}{*}{\makecell{Remove \\ Clamp}} & 
    \multicolumn{2}{c|}{Few Gaussians} & 
    \multicolumn{2}{c}{Many Gaussians} \\
    & & & PSNR$\uparrow$ & EF-PSNR$\uparrow$ & PSNR$\uparrow$ & EF-PSNR$\uparrow$ \\
    \midrule
     &  &  & 27.10 & 22.73 & 33.41 & \underline{30.33} \\
     &  & \checkmark & \underline{27.29} & 22.92 & \textbf{33.57} & \textbf{30.48} \\
    \checkmark &  &  & 26.98 & 22.72 & 33.25 & 29.36 \\
    \checkmark &  & \checkmark & 27.15 & 22.92 & 33.20 & 29.31 \\
    \checkmark & \checkmark &  & 27.06 & \underline{23.09} & 33.28 & 29.44 \\
    \checkmark & \checkmark & \checkmark & \textbf{27.31} & \textbf{23.41} & \underline{33.44} & 29.18 \\
    \bottomrule
  \end{tabular}
\end{table}

\section{Conclusion}
    In this work, we presented a contour-guided 2DGS framework that explicitly incorporates structural information into the rasterization process.
    Our design integrates three key components: contour guidance, warm-up training, and removal of the clamping, which work in a complementary manner.
    Experiments on synthetic charts and the DAVIS dataset demonstrate that our method improves reconstruction quality, particularly around edges and under constrained Gaussian budgets, achieving higher PSNR and EF-PSNR with sharper boundaries and richer intermediate colors.
    While the benefits are most significant when the number of Gaussians is limited, our method also remains competitive in high-capacity settings, indicating a strong generalization across different reconstruction regimes.
    These findings confirm that explicit structural guidance is an effective principle for improving Gaussian-based representations.
    Future work includes extending our framework to dynamic scenes, incorporating semantic or instance-level guidance, and exploring applications in video compression and interactive image editing.

\bibliography{reference}
\bibliographystyle{IEEEtran}

\vspace{12pt}

\end{document}